\documentclass[]{bytedance_seed}

\usepackage[toc,page,header]{appendix}

\usepackage{amsmath}
\usepackage{amssymb}
\usepackage{hyperref}
\usepackage[most]{tcolorbox}
\usepackage{listings}
\usepackage{xcolor}
\usepackage{xspace}
\usepackage{listings}
\usepackage{wrapfig}
\DeclareUnicodeCharacter{00FC}{{\rmfamily\"u}}
\definecolor{jsonkey}{rgb}{0.36,0.54,0.66}
\definecolor{jsonstring}{rgb}{0.58,0.36,0.18}
\definecolor{jsonnumber}{rgb}{0.2,0.2,0.2}

\lstdefinelanguage{json}{
  basicstyle=\ttfamily\footnotesize,
  showstringspaces=false,
  breaklines=true,
  frame=none,
  string=[b]",
  stringstyle=\color{jsonstring},
  keywordstyle=\color{jsonkey},
  morekeywords={true,false,null},
  literate=
   *{0}{{\color{jsonnumber}0}}1
    {1}{{\color{jsonnumber}1}}1
    {2}{{\color{jsonnumber}2}}1
    {3}{{\color{jsonnumber}3}}1
    {4}{{\color{jsonnumber}4}}1
    {5}{{\color{jsonnumber}5}}1
    {6}{{\color{jsonnumber}6}}1
    {7}{{\color{jsonnumber}7}}1
    {8}{{\color{jsonnumber}8}}1
    {9}{{\color{jsonnumber}9}}1
}

\newcommand{\NAME}{{ActWorld}\xspace}

\title{\NAME: From Explorable to Interactive World Model via Action-Aware Memory}

\author[1,2,*]{Zhexiao Xiong}
\author[2]{Yizhi Song}
\author[2]{Hao Kang}

\author[2]{Qing Yan}
\author[2]{Liming Jiang}
\author[2]{Jenson Yang}
\author[2]{Zhoujie Fu}
\author[2]{Stathi Fotiadis}
\author[2]{Angtian Wang}
\author[2]{Zichuan Liu}
\author[2]{Bo Liu}
\author[2]{Yiding Yang}
\author[2]{Xin Lu}
\author[1]{Nathan Jacobs}

\affiliation[1]{Washington University in St. Louis}
\affiliation[2]{Intelligent Creation, ByteDance}

\contribution[*]{Work done during internship at ByteDance}

\abstract{
Interactive world models aim to simulate environment dynamics under real-time user actions. However, their action vocabulary is largely confined to navigation: most actions correspond to motion (e.g., walk, turn, look around), while interaction with objects in the scene (e.g., pick up plates, open doors, or trigger physical responses) is either absent, restricted to game domains, or relegated to prompt-to-full-video scenarios.
The resulting worlds are visually explorable but not truly
actionable.
In this work, we present \textbf{\NAME}, an interactive world model
that extends prior navigation-centric generators to support
mid-rollout object interaction within a chunk-autoregressive
framework.
We argue that the navigation--interaction gap stems from two bottlenecks. First, a data bottleneck: the lack of human--object interaction data with accurate, dense labels. Second, a memory bottleneck: recency-biased history compression in existing world models discards the event-transition frames that causally determine subsequent object states, leading to an action-forgetting pathology.
On the data side, we construct a \textbf{100K interaction video dataset}, each annotated with per-chunk captions via chain-of-thought reasoning. On the model side, we introduce a \textbf{hierarchical action-aware memory} design that routes history compression by interaction importance, complemented by a persistent memory bank that maintains event-update and object-identity tokens across long rollouts.
Experiments show that \NAME supports both flexible navigation and rich object interaction within a single model, substantially improving
interaction fidelity over navigation-only baselines without sacrificing
viewpoint control. Project page is available at \url{https://interactwm.github.io/ActWorld}.
}

\date{May 22, 2026}

\begin{document}
\maketitle
\begin{figure}[ht]
\vspace{-25pt}
    \centering
    \includegraphics[width=\linewidth]{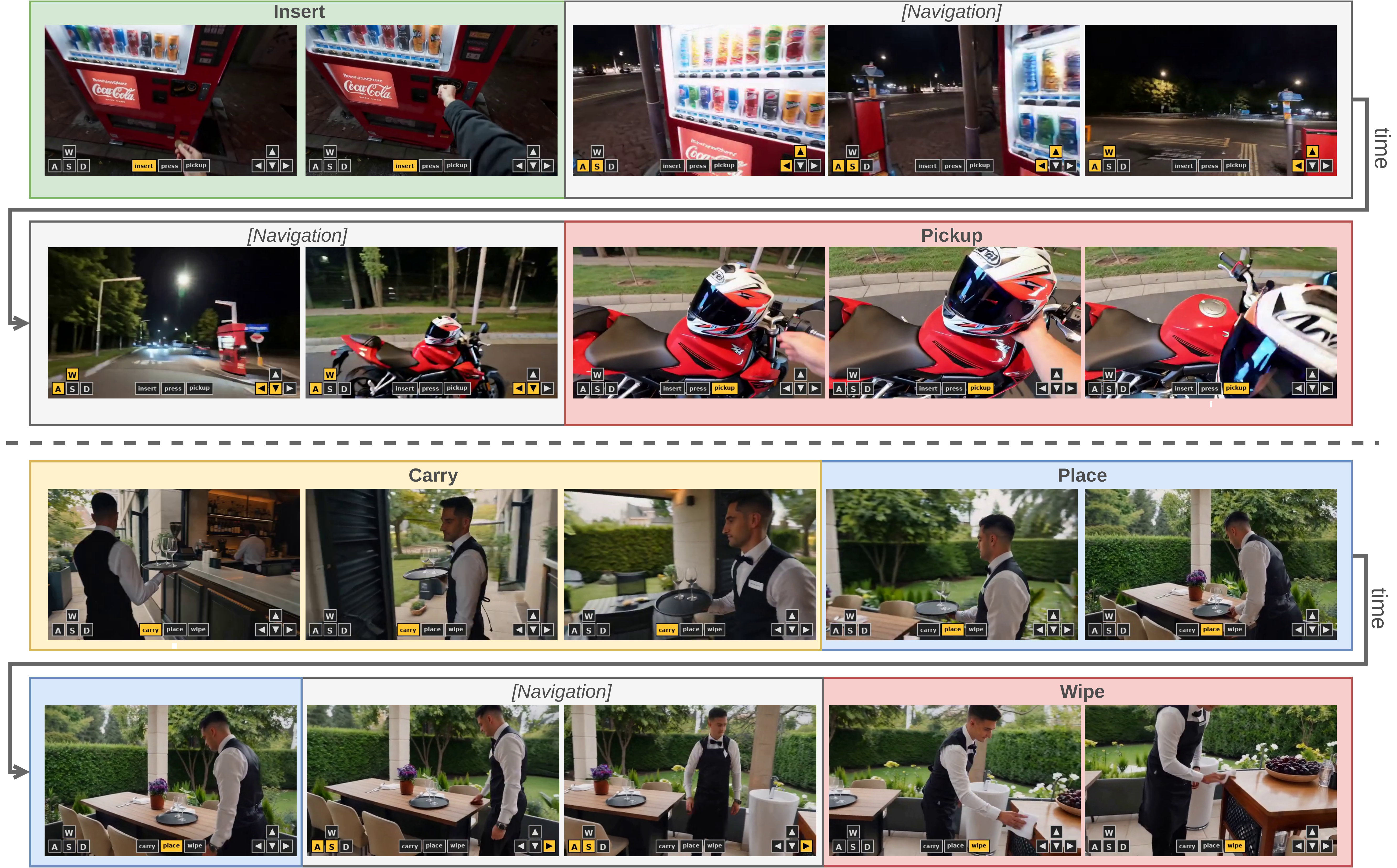}
    \caption{\textbf{\NAME} is an interactive world model that handles both long-horizon navigation and mid-rollout object
    interaction within a single rollout, under per-frame keyboard and
    mouse control (WASD~+~arrow keys overlaid on each frame; yellow marks
    the active key). Each row is one continuous trajectory: time flows
    left-to-right and wraps into the next row (arrow $t$); colored bands
    separate navigation segments from object-interaction
    segments, with the action label shown above.}
    \label{fig:teaser}
\end{figure}


\section{Introduction}

Building world models that can simulate environment dynamics and support real-time interaction has emerged as a central challenge in artificial intelligence, with broad implications for gaming, embodied AI, autonomous driving, and content creation. The long-term goal is to learn how the world evolves directly from visual experience and to allow agents, whether human or artificial, to act within the simulated environment through controllable inputs.
Early efforts~\cite{valevski2024diffusion,parkerholder2024genie2} established the feasibility of neural world simulation with diffusion-based generators. Since then, advances has been made on long-horizon consistency~\cite{wu2026infinite,2026matrix}, real-time high-resolution generation~\cite{worldplay2025,hyworld2025,wang2026worldcompass,2026matrix,zhao2025taste}, and cross-domain controllability~\cite{lingbot-world,mao2025yume15,xiang2025pan}.

Despite this progress, a fundamental limitation remains: \textbf{existing interactive world models predominantly focus on locomotion and viewpoint control, while largely neglecting object-level interaction}. The majority of current systems~\cite{worldplay2025,2026matrix,wu2026infiniteworld,mao2025yume15,lingbot-world} condition generation on keyboard or mouse inputs that drive movement and camera changes, producing worlds that are visually explorable but not truly actionable---agents can walk, turn, and navigate, but cannot pick up items, open doors, manipulate tools, or trigger meaningful physical responses from the environment. While PAN~\cite{xiang2025pan} supports language-specified manipulation commands, it operates in an offline, non-real-time regime unsuitable for interactive exploration. Domain-specific models such as Solaris~\cite{solaris2026} and Matrix-Game~\cite{zhang2025matrixgame,he2025matrix} do support rudimentary object interactions (e.g., block breaking in Minecraft), they are confined to simple game environments and suffer from significant quality degradation during complex action sequences. More critically, no existing method jointly supports both fine-grained object interaction and flexible locomotion control within a unified real-time framework.

In this work, we present \textbf{\NAME}, an interactive world model that unifies \textbf{rich object interaction} and \textbf{flexible viewpoint control} within a single real-time, chunk-autoregressive framework.
We argue that the navigation--interaction gap stems from two compounding bottlenecks.
The first is data: existing world-model datasets are overwhelmingly navigation-centric, providing little supervision for object-level dynamics.
The second is memory design: current models organize history simply by temporal recency, preserving recent observations while compressing older ones. So frames where object states have been modified (e.g., a bottle being filled, a lamp turned on) are often heavily compressed when they are distant from the current video chunk. As a result, the model may lose the necessary evidence to predict subsequent states.
We address both: on the data side, we collect a 100K interaction-dense video dataset with per-chunk dense captions from proprietary models; on the model side, we align memory granularity with interaction events, rather than time alone, so that causally critical frames survive compression regardless of their age. Our contributions are as follows:

\begin{itemize}
\item \textbf{Hierarchical action-aware memory}. To address the action-forgetting behavior, we introduce a novel memory mechanism: A local memory bank routes and amplifies interaction-critical frames within the sliding window; a persistent memory bank maintains compact event-update and object-identity tokens that survive beyond the window's eviction horizon.

\item \textbf{An interaction-dense dataset and annotation pipeline}. Existing world-model datasets are largely navigation-centric, lacking both object-interaction coverage and fine-grained temporal labels. We construct a high-quality 100K-video dataset using proprietary video models, spanning 40 action categories, and annotate every chunk with a dense caption and an interaction-phase label.

\item \textbf{A real-time interactive world model with navigation and object interaction}. We propose \NAME, a real-time interactive world model that jointly supports flexible navigation and rich object interaction within a single framework. \NAME integrates action-aware memory and the interaction data pipeline with a dual-branch camera conditioning module. Our model is validated on \textit{I-Bench}, a new long-horizon benchmark that interleaves navigation and object interaction, where \NAME substantially outperforms existing world models.

\end{itemize}

Experiments show that \NAME supports flexible navigation and rich object interaction within a single real-time model. Compared with navigation-centric baselines, it substantially improves interaction fidelity while preserving locomotion and viewpoint controllability.

\section{Related Works}

\textbf{Long Video Generation.}
Most modern video diffusion models~\cite{wan2025,kong2024hunyuanvideo} use bidirectional attention across all frames, which precludes streaming generation and real-time interaction. Causal autoregressive variants avoid this but suffer from exposure bias, where errors compound as the model conditions on its own imperfect outputs. A line of recent work closes this train--test gap through distillation or rollout-aware training: asymmetric distillation from a bidirectional teacher~\cite{yin2025causvid}, video-level supervision over self-generated sequences~\cite{huang2025selfforcing}, autoregressive-teacher ODE initialization~\cite{zhu2026causal}, and teacher-free training with injected history noise~\cite{guo2025endtoendtrainingautoregressivevideo}, together making single-GPU streaming generation practical. These methods, however, target open-domain text-to-video synthesis and do not address action conditioning, spatial memory, or object-level interaction.

\textbf{Real-Time Video Generation.}
Even with causal architectures, multi-step sampling remains the main throughput bottleneck. Distribution Matching Distillation~\cite{yin2024onestep,yin2024improved} and consistency distillation~\cite{song2023consistency,luo2023lcm} compress many-step teachers into few-step students, with CausVid~\cite{yin2025causvid} extending DMD to causal video generation. Engineering-level optimizations such as INT8 attention, sequence parallelism, and lightweight VAE decoder distillation~\cite{2026matrix,hyworld2025} further cut per-frame latency, and combined pipelines now reach 24--40\,FPS at 720p~\cite{worldplay2025,2026matrix}. These accelerations are designed for unconditional or text-conditioned generation; preserving fidelity under simultaneous locomotion and object-interaction conditioning remains underexplored.

\textbf{World Models.}
World models learn environment dynamics from data and simulate them in response to actions, with substantial progress across autonomous driving~\cite{russell2025gaia,xiong2026unidrive,liu2026towards,li2024enhancing} and robotics~\citep{cen2025worldvla,zhu2025unified,jiang2026wovr}. Existing approaches are broadly 3D-based, synthesizing explorable scene geometry~\cite{xiong2025panodreamer,yu2024wonderjourney,hunyuanworld2025tencent,hyworld22026}, or video-based, generating future observations directly as video. Early video-based systems established feasibility in constrained, single-domain settings~\citep{valevski2024diffusion,parkerholder2024genie2,oasis2024} but suffered from short horizons and fragile spatial memory. More recent work advances along distinct axes: Infinite-World~\cite{wu2026infinite} and Matrix-Game~3.0~\cite{2026matrix} extend coherent generation beyond $1{,}000$ frames through memory compression and self-correction; WorldPlay~\cite{worldplay2025,hyworld2025,wang2026worldcompass} and Matrix-Game~3.0 reach real-time frame rates at high resolution; LingBot-World~\cite{lingbot-world} and Yume-1.5~\cite{mao2025yume15} generalize across visual domains; and a further line improves the physical plausibility of generated dynamics~\cite{agarwal2026cosmos,xiong2026physalign,liu2024physgen}. However, action conditioning remains dominated by locomotion and viewpoint changes; object-level interaction is either absent, confined to narrow game manipulations~\cite{oasis2024,zhang2025matrixgame,he2025matrix}, relegated to offline language-conditioned generation~\cite{xiang2025pan}, or restricted to static, grounded image-level control~\cite{groundingbooth,wang2025ms}. \textbf{\NAME} departs from this navigation-centric paradigm by jointly addressing the dataset scarcity and memory-design limitations that result in the navigation--interaction gap.

\section{Method}

\begin{figure*}[t!]
    \centering
    \includegraphics[width=\linewidth]{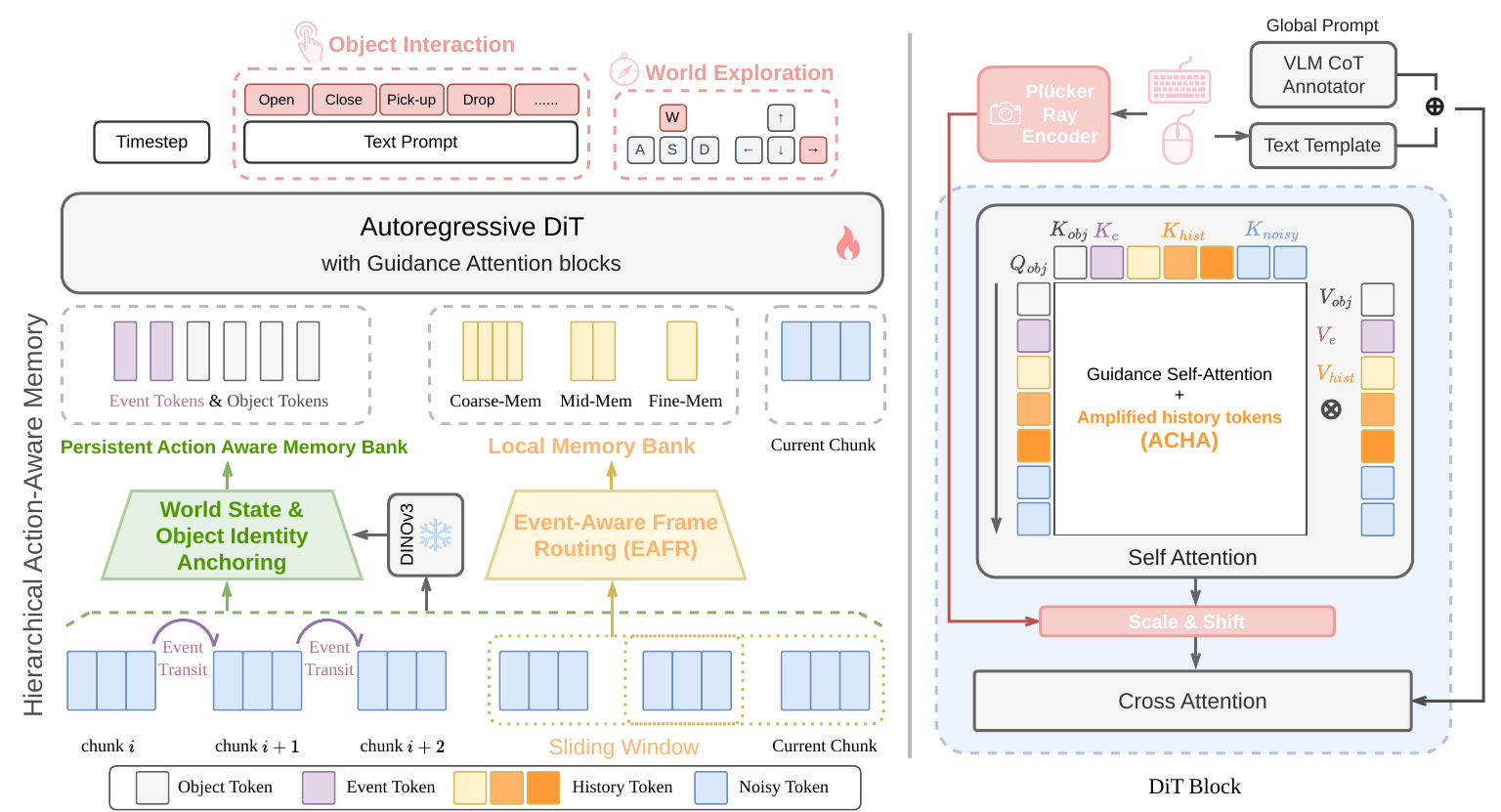}

\caption{
\textbf{\NAME} \textbf{pipeline.} An autoregressive DiT generates each chunk under high-level object-interaction commands and low-level keyboard/mouse controls. \textbf{Left:} past observations flow through a hierarchical action-aware memory with two channels---a persistent action aware bank of event/object tokens that survives long navigation gaps, and a local bank of EAFR-routed coarse/mid/fine history tokens. \textbf{Right:} inside each DiT block, self-attention amplifies history keys via ACHA, Plücker rays drive a per-token scale-and-shift, and cross-attention ingests the per-chunk caption combined with a symbolic text-camera embedding.
}
    \label{fig:pipeline}
\end{figure*}

\NAME generates videos in a chunk-autoregressive manner, where each chunk is conditioned on past observations, user actions, and camera controls. Our method has three main components: per-chunk semantic captions for localized language conditioning (\S\ref{sec:cot_caption}), decoupled camera control through Plücker-ray FiLM and symbolic text-camera channels (\S\ref{sec:method_kmctrl}), and a hierarchical action-aware memory pipeline that re-routes interaction-relevant frames and persists event/object anchors across the latent buffer's eviction horizon (\S\ref{sec:method_highlevel_action}). A short distillation stage further reduces sampling cost for real-time inference (\S\ref{sec:method_distill}). We present the whole pipeline in Fig.~\ref{fig:pipeline}.

\subsection{Chain-of-Thought Per-Chunk Annotation}
\label{sec:cot_caption}

A single video-level caption is too coarse for chunk-autoregressive generation. Since the same caption is shared by multiple visually distinct chunks, the text condition does not specify which interaction phase the current chunk should depict, leading to temporally ambiguous conditioning and repetitive content. We instead annotate each chunk offline with a dedicated description and the structured labels consumed by \S\ref{sec:method_highlevel_action}. Every video is divided into non-overlapping 33-frame segments ($\approx$1.4\,s at 24\,fps, aligned with the VAE latent window). For each chunk we extract 5 evenly-spaced keyframes and query a vision-language model (GPT-5.4). A direct prompt (``describe this chunk'') yields hallucinated interactions, generic ``scene continues'' fallbacks, or confusion between camera motion and object interaction; we find that chain-of-thought (CoT) prompting substantially improves quality by forcing the VLM to reason over explicit visual evidence before committing to its outputs. The prompt instructs the model to:

\begin{itemize}\setlength{\itemsep}{0pt}\setlength{\parskip}{0pt}
  \item compare consecutive keyframe pairs and enumerate observable changes (object displacements, new contacts such as a hand grasping an object or an object placed on a surface, state transitions), ignoring camera motion;
  \item synthesize these observations to decide whether active interaction is occurring ($y^{\text{int}}_k$) and classify its phase $y^{\text{ph}}_k$ into one of six phases: \textit{approaching}, \textit{reaching}, \textit{contact}, \textit{manipulating}, \textit{completing}, and \textit{post-action}. These phases respectively denote movement toward the target, object-directed reach, first physical touch, sustained object motion, final release or settlement, and frames after the action has finished;
  \item emit a 1--2 sentence description grounded solely in the accumulated frame-level evidence.
\end{itemize}
To maintain temporal coherence, chunks within each video are annotated sequentially, with chunk $t{-}1$'s description provided as context when annotating chunk~$t$. The per-chunk descriptions are encoded offline by the frozen UMT5~\citep{chung2023unimax} text encoder into embeddings $\mathbf{e}_t^{\mathrm{chunk}} \in \mathbb{R}^{L \times d}$ used as the cross-attention condition; the structured labels $(y^{\text{int}}_k, y^{\text{ph}}_k)$ are passed downstream to the memory components in \S\ref{sec:method_highlevel_action}.

\subsection{Keyboard/Mouse Control Conditioning}
\label{sec:method_kmctrl}

We distinguish low-level keyboard/mouse controls (camera translations and rotations) from high-level actions handled in \S\ref{sec:method_highlevel_action}, and expose the former through two complementary branches.

\paragraph{Geometric branch.}
Per-frame camera trajectories are converted into a per-pixel Plücker-ray tensor and routed through a shared \texttt{PlückerFiLM} module, following~\citep{lingbot-world}. Our only departure is to reuse the module weights across all transformer blocks (parameter overhead $\approx$1.2\% of the 14B DiT). The module emits per-token, zero-initialised FiLM scale-and-shift signals applied only to current-chunk hidden states (history KV is untouched), so toggling the branch off on a pretrained checkpoint recovers the baseline exactly.

\paragraph{Symbolic text-camera branch.}
The Plücker branch encodes camera motion as continuous geometry, but supervision is given as discrete user commands ($9{\times}9{=}81$ \texttt{(keyboard,\,mouse)} combinations such as \texttt{W+D} with $\uparrow{+}\rightarrow$). Following~\citep{mao2025yume15}, we map each command to a short natural-language template, encode it once with the frozen UMT5 text encoder, cache the embedding, and concatenate it with the per-chunk caption (\S\ref{sec:cot_caption}) on the cross-attention path. An independent dropout $p_{\text{cam-txt}}{=}0.1$ on this branch discourages shortcut learning and biases the model toward the geometric signal.

Plücker-ray construction, pose normalisation, full FiLM equations, the 81-entry command vocabulary, and ablations are in Appendix.~\ref{sec:supp_kmctrl}.

\subsection{Hierarchical Action-Aware Memory for High-Level Interactions}
\label{sec:method_highlevel_action}

The conditioning mechanisms introduced so far control what each chunk depicts (\S\ref{sec:cot_caption}) and how the camera moves (\S\ref{sec:method_kmctrl}), but neither shapes which past frames the model conditions on. This third axis becomes critical under interaction commands, where the next chunk's outcome often hinges on sparse \textit{contact} or \textit{manipulating} frames many steps back. We diagnose this memory-side gap below and address it without modifying the backbone.

\paragraph{Problem formulation: action-forgetting in recency-based memory.}
Our backbone organizes the history buffer $\mathcal{H}=\{h_1,\dots,h_{t-1}\}$ via a Multi-Term Memory (MTM) scheme that partitions past chunks into three buckets $\mathcal{S}_{\text{short}}, \mathcal{S}_{\text{mid}}, \mathcal{S}_{\text{long}}$ (named after their temporal distance) by recency, patchified with progressively larger spatio-temporal kernels and concatenated as history keys/values for DiT self-attention. For roaming trajectories this is nearly optimal: the frames that constrain the next observation are overwhelmingly recent. Interactive generation breaks this assumption. The frames that causally determine the next observation under one of our 40 high-level action commands (e.g., \texttt{pickup}, \texttt{open}, \texttt{attach}; full list in Appendix.~\ref{sec:supp_actions}) are not recent ones but the sparse \textit{contact} and \textit{manipulating} moments, which typically sit in $\mathcal{S}_{\text{mid}}$ or $\mathcal{S}_{\text{long}}$ where object pose and contact geometry are degraded most; worse, the buffer's finite reach evicts anything older than $|\mathcal{H}|$ chunks, so objects that briefly leave the frame are lost on re-visit. We call this the action-forgetting problem: it is a memory-design rather than capacity issue, because \textit{contact} and \textit{manipulating} frames remain coarsely patchified no matter how far back the buffer reaches or how large the backbone grows.

Our response keeps the backbone unchanged and adds two complementary stages: a memory-retrieval re-routing stage that re-shapes which past frames the existing buckets, and a persistent action-aware memory bank that survives the latent buffer's eviction horizon. 
Both share three per-chunk labels emitted once offline by the CoT annotator of \S\ref{sec:cot_caption}---an interaction flag $y^{\text{int}}_k$, a phase $y^{\text{ph}}_k$, and a video-level action class $a$---and are designed so that toggling them off recovers the baseline bit-for-bit.

\paragraph{Event-aware frame re-assignment (EAFR).}
\label{sec:mem_retrieval}
We replace MTM's time-based bucketing with an importance-ranked split. For each past chunk $k$ we score
\begin{equation}
  w_k \;=\; \lambda_\varphi\,\varphi\!\left(y^{\text{int}}_k,\,y^{\text{ph}}_k\right)
       \;+\; \lambda_r\,\exp\!\left(-(t-k)/\tau\right),
  \label{eq:eafr}
\end{equation}
where $\varphi$ is a fixed phase prior that peaks on \textit{contact} and \textit{manipulating} and decays elsewhere, $\lambda_\varphi, \lambda_r > 0$ are scalar mixing weights, and the second term retains a recency bias with timescale $\tau$ relative to the current chunk index $t$. The buckets are filled greedily by descending $w_k$ under the backbone's original size budgets, so we hereafter refer to them as $\mathcal{S}_{\text{fine}}, \mathcal{S}_{\text{mid}}, \mathcal{S}_{\text{coarse}}$. A \textit{contact} frame from many steps ago can sit in fine-grained $\mathcal{S}_{\text{fine}}$ while a purely navigational recent frame drops to $\mathcal{S}_{\text{coarse}}$. RoPE positions are preserved across re-assignment: a migrated frame keeps its original time index, so EAFR changes which compressor a frame sees without altering its temporal position.

\paragraph{Action-conditioned history amplification (ACHA).}
The backbone already exposes a learnable per-head amplifier $s\!\in\!\mathbb{R}^{n_h}$ applied to history keys in self-attention, identical by default for a reach-to-grasp clip and a walk-past clip. ACHA makes it action-conditioned:
\begin{equation}
  \alpha(\mathbf{e}_a)
  \;=\; \operatorname{softplus}\!\Big(s + W_2\,\sigma(W_1\,\mathbf{e}_a)\Big),
  \qquad
  K'_{\text{hist}} = \alpha(\mathbf{e}_a)\odot K_{\text{hist}},
  \label{eq:acha}
\end{equation}
where $\mathbf{e}_a\!\in\!\mathbb{R}^{d}$ is a learned embedding of the action class $a$, $\sigma$ is SiLU, and $W_1, W_2$ form a $d\!\to\!d_b\!\to\!n_h$ bottleneck ($d_b{=}256$). The last layer is zero-initialised, so $\alpha$ collapses to the baseline amplifier at step~0; during training the MLP learns to sharpen attention onto history keys causally relevant to the current action, without adding tokens or a new attention stream.

\paragraph{Persistent action-aware memory bank.}
\label{sec:mem_bank}
EAFR and ACHA still operate on the pixel-space latent stream, which is an inefficient carrier for sparse symbolic facts (``an object just left the gripper'') and is bounded by the latent buffer's eviction horizon (anything older than $|\mathcal{H}|$ chunks is gone). We add a small structured channel that survives both limitations: an action-aware memory bank of at most $K_{\text{tot}}$ tokens (default $K_{\text{tot}}{=}16$), preserved across chunk boundaries under a FIFO policy with phase-driven pinning, and prepended to the DiT self-attention input. The bank carries two complementary kinds of tokens: event tokens that record what happened and when, and object tokens that record what the affected object looks like. Both share the bank's segment slot and FIFO policy, and differ only in what triggers them and what they encode.

\textbf{\textit{Event tokens.}} These fire at phase transitions: boundary events derived from changes in $y^{\text{ph}}$ between consecutive chunks, complementing the per-chunk phase label itself. A rule-based writer scans $y^{\text{ph}}$ for three such transitions---\textsc{Enter-Manip} (onset of \textit{manipulating}), \textsc{Enter-Complete} (onset of \textit{completing}), and \textsc{Release} (end of \textit{manipulating})---and emits one token per firing at chunk~$k$:
\begin{equation}
  \mathbf{e}^{\text{evt}}_k \;=\;
     E_\xi[\xi_k] + E_{\text{ph}}[y^{\text{ph}}_k] + E_a[a]
     + \operatorname{AttnPool}(\mathbf{h}_k),
  \label{eq:evt}
\end{equation}
where $\xi_k$ is the firing transition tag; $E_\xi, E_{\text{ph}}, E_a$ are learned embedding tables for the transition, phase, and action class; their labels reuse the same vocabulary as EAFR/ACHA and $E_a$ shares parameters with ACHA's action embedding, so no new vocabulary is introduced. $\operatorname{AttnPool}(\mathbf{h}_k)$ is a single-query attention pool over the triggering chunk's patchified latent tokens, providing a visual summary of what the model just generated at the transition. Recency is left to the bank's FIFO ordering rather than encoded explicitly, so these tokens index memory by transition-tag $\times$ phase.

\textbf{\textit{Object tokens.}} These capture the visual identity of the touched region. Whenever a chunk has $y^{\text{int}}_k\!=\!1$ or carries a phase in $\{\textit{completing}, \textit{post-action}\}$, we emit up to $K_{\text{pc}}$ object-anchor tokens (default $K_{\text{pc}}{=}3$) drawn from a frozen DINOv3 encoder. Concretely, for each of $K_{\text{kf}}{=}5$ evenly-spaced keyframes of chunk~$k$ we run DINOv3 to obtain patch features, score every patch by its $L_2$ norm (a saliency proxy that suppresses uniform-background patches), and keep the top $K_{\text{pc}}$ across the chunk. Each surviving patch $\mathbf{f}^{\text{dino}}_{k,j}\!\in\!\mathbb{R}^{768}$ becomes a token:
\begin{equation}
  \mathbf{e}^{\text{obj}}_{k,j} \;=\;
     W_v\!\bigl(\mathbf{f}^{\text{dino}}_{k,j}\bigr)
     + E_{\text{ph}}[y^{\text{ph}}_k] + E_a[a],
  \label{eq:obj}
\end{equation}
where $W_v$ is a two-layer MLP with a zero-initialised final layer that lifts the patch feature into the transformer width $d$; this zero-init means the bank emits the zero vector at step~0, so toggling the writer on a pretrained checkpoint preserves the baseline forward pass exactly until learning kicks in. The phase and action embeddings $E_{\text{ph}}, E_a$ are reused from event tokens; what makes object tokens distinct is the $W_v(\mathbf{f}^{\text{dino}})$ term, which encodes the visual signature of the touched region rather than a transition tag. As with event tokens, recency is left to FIFO ordering, so these tokens index memory by visual signature $\times$ phase. Because DINOv3 features are view-stable, an anchor written at chunk~$k$ remains a useful query target after the camera turns away and back, letting the bank carry an object identity across long navigation gaps.

\textbf{\textit{Bank operation and inference cost.}} 
Tokens enter a FIFO deque of capacity $K_{\text{tot}}{=}16$, except those whose source chunk has $y^{\text{ph}}\in\{\textit{contact}, \textit{manipulating}, \textit{completing}\}$, which are pinned for the rollout, preserving interaction moments through long stretches of navigation. The bank is prepended to the DiT input and enters self-attention only. At inference, object anchors are computed online by VAE-decoding each generated chunk and running a frozen DINOv3~\citep{simeoni2025dinov3} forward; gating this on interaction-related chunks keeps the average overhead below 15\%. Bank mechanics (RoPE handling, segment embedding, training-time anchor pre-baking) are deferred to the appendix.

\subsection{Few-Step Distillation for Long-Horizon Action-Conditioned Generation}
\label{sec:method_distill}

For interactive use---each chunk must land in a fraction of a second---we follow the Stage-1 chunk-AR training above with two further stages from the Helios~\cite{yuan2026helios} acceleration recipe: (i) a multi-resolution flow-matching stage that splits denoising into $K{=}3$ resolution levels and regresses on the linear-flow velocity $\mathbf{v}^{k}=\mathbf{x}^{k}-\mathrm{Upsample}(\mathbf{x}^{k-1})$, processing fewer tokens at coarser levels; and (ii) an adversarial DMD-style distillation that reduces the 50-step teacher to a 3-step generator $G_\theta$. All conditioning streams from \S\ref{sec:cot_caption}--\ref{sec:method_highlevel_action} ride through both stages as DiT inputs unchanged; full equations and hyperparameters follow~\cite{yuan2026helios}.

\paragraph{I2V ODE warm-up.}
Our one deviation from~\cite{yuan2026helios} is the ODE-pair warm-up. Rather than t2v pairs, we construct pairs in an image-to-video regime: each clip's first ground-truth chunk is fixed as a clean conditioning image, and the teacher runs the multi-step sampler only on subsequent chunks under the same action labels and Plücker rays the student will later see. This matches the streaming-i2v interface at deployment (the user always provides a starting frame), removes regression variance from an extra t2v denoising step, and converges noticeably faster than t2v pairs of comparable size.

After distillation each 33-frame chunk is produced in three sampling steps without classifier-free guidance; the action-text dropout from \S\ref{sec:method_kmctrl} is kept on throughout Stage-3 so the CFG-free student still reacts to changing keyboard/mouse inputs. Full distillation hyperparameters are in Appendix.~\ref{sec:supp_distill}.

\section{Experiments}

\subsection{Data Generation Pipeline}
Our training data combines two complementary sources. (1) We synthesize 100K videos with proprietary models to provide dense human-motion supervision, comprising 55K first-person and 45K third-person clips spanning 40 action categories; the third-person split covers both human-centric and non-human-centric actions. (2) We incorporate 400 hours of real-world data from a public dataset, mainly consisting of egocentric walking sequences in natural environments together with game footage. Detailed statistics of the data composition are provided in the appendix.

\subsection{I-Bench: Long-Horizon Action--Navigation Benchmark}
\label{sec:ibench}

Existing benchmarks isolate either roaming through passive scenes (e.g., Yume, Sekai, VBench-Long) or scripted object interactions with fixed camera (WorldModelBench, VideoPhy-2); neither stresses what an interactive world model faces at deployment: chaining interactions while translating the viewpoint. We introduce \textbf{I-Bench}, in which every clip executes a long-horizon \textbf{(action sequence, camera sequence)} composite script.

I-Bench contains $300$ prompts evenly split between first- and third-person views, organized as $30$ semantically coherent sequences of $10$ prompts each. Each prompt composes three action verbs from a $40$-verb vocabulary interleaved with $2$--$3$ camera primitives (translations, pans, tilts) described only in natural language, so the model must recover both signals from the prompt. Clips span $10$ chunks of $33$ frames. Each clip is annotated with a global caption, per-chunk phase and sub-action descriptions, and a VIPE-recovered camera trajectory~\citep{huang2025vipe}; details are shown in the appendix.

\begin{table*}[t]
\centering
\footnotesize
\renewcommand{\arraystretch}{0.9}
\setlength{\tabcolsep}{6pt}
\caption{\textbf{VBench perceptual and consistency metrics on I-Bench.} 
SC: subject-consistency, BC: background-consistency, MS: motion-smoothness, 
AQ: aesthetic-quality, IQ: imaging-quality, DD: dynamic-degree, 
TF: temporal-flickering, OC: overall-consistency, i2v-S / i2v-B: VBench-i2v subject / background. 
All scores are higher-is-better; bold marks the column best.}
\label{tab:vbench}
\begin{tabular}{l|cccccccc|cc}
\toprule
Method & SC & BC & MS & AQ & IQ & DD & TF & OC & i2v-S & i2v-B \\
\midrule
Yume 1.5~\cite{mao2025yume15} & 0.743 & 0.872 & 0.985 & 0.432 & 0.645 & 1.000 & 0.962 & 0.180 & 0.930 & 0.943 \\
HY-World 1.5~\cite{hyworld2025} & 0.856 & 0.873 & 0.993 & 0.449 & 0.734 & 0.933 & \textbf{0.977} & 0.199 & 0.952 & 0.953 \\
Lingbot-World~\cite{lingbot-world} & 0.724 & 0.866 & 0.979 & 0.436 & 0.693 & 0.983 & 0.966 & 0.182 & 0.917 & 0.935 \\
Matrix-Game 3~\cite{2026matrix} & 0.662 & 0.854 & 0.981 & 0.377 & 0.681 & 1.000 & 0.954 & 0.169 & 0.862 & 0.889 \\
Astra~\cite{zhu2025astra} & 0.603 & 0.841 & 0.946 & 0.290 & 0.465 & 0.817 & 0.946 & 0.152 & 0.759 & 0.827 \\
Infinite-World~\cite{wu2026infinite} & 0.801 & 0.863 & 0.987 & 0.442 & \textbf{0.748} & 0.967 & 0.959 & 0.087 & 0.716 & 0.743 \\
\midrule
\textbf{Ours} & \textbf{0.871} & \textbf{0.896} & \textbf{0.991} & \textbf{0.485} & 0.731 & \textbf{1.000} & 0.973 & \textbf{0.201} & \textbf{0.954} & \textbf{0.957} \\
\bottomrule
\end{tabular}
\end{table*}

\begin{table*}[t]
\centering
\footnotesize
\renewcommand{\arraystretch}{0.9}
\begin{minipage}[t]{0.48\linewidth}
\centering
\setlength{\tabcolsep}{6pt}
\caption{\textbf{VLM-AJ (VLM-Action-Judge) on I-Bench.}
$\text{IF}$: mean instruction-following score (0--3);
\textsc{Succ.}: success rate (Level\,3); $\geq\!2$: partial-or-better
rate (Level\,$\geq 2$).  Higher is better.}
\label{tab:vlm_aj}
\begin{tabular}{l | ccc}
\toprule
Method & $\text{IF}\uparrow$ & Succ.\,$\uparrow$ & $\geq\!2\uparrow$ \\
\midrule
Yume 1.5~\cite{mao2025yume15} & 1.638 & 20.12 & 46.75 \\
HY-World 1.5~\cite{hyworld2025} & 0.709 & 5.19 & 18.70 \\
Lingbot-World~\cite{lingbot-world} & 1.635 & 19.89 & 49.91 \\
Matrix-Game 3~\cite{2026matrix} & 0.295 & 1.88 & 8.27 \\
Astra~\cite{zhu2025astra} & 0.949 & 5.83 & 19.36 \\
Infinite-World~\cite{wu2026infinite} & 0.237 & 1.51 & 3.95 \\
\midrule
\textbf{Ours} & \textbf{2.557} & \textbf{57.8} & \textbf{84.5} \\
\bottomrule
\end{tabular}
\end{minipage}%
\hfill
\begin{minipage}[t]{0.48\linewidth}
\centering
\setlength{\tabcolsep}{5pt}
\caption{\textbf{KMF (Key-Mouse-Following) on I-Bench.} $\mathrm{Acc}_{\text{full}}$: joint (keys, mouse) match; $\mathrm{Acc}_{\text{keys}}$ / $\mathrm{Acc}_{\text{mouse}}$: per-axis accuracy. Computed from VIPE-recovered $\mathrm{SE}(3)$ trajectories.}
\label{tab:kmf}
\begin{tabular}{l | ccc}
\toprule
Method & $\mathrm{Acc}_\text{full}\uparrow$ & $\mathrm{Acc}_\text{keys}\uparrow$ & $\mathrm{Acc}_\text{mouse}\uparrow$ \\
\midrule
Yume 1.5~\cite{mao2025yume15} & 4.82 & 31.79 & 15.71 \\
HY-World 1.5~\cite{hyworld2025} & 9.17 & 42.78 & 25.56 \\
Lingbot-World~\cite{lingbot-world} & 2.67 & 28.00 & 13.33 \\
Matrix-Game 3~\cite{2026matrix} & 20.00 & \textbf{45.01} & 40.83 \\
Astra~\cite{zhu2025astra} & 11.43 & 28.57 & 28.57 \\
Infinite-World~\cite{wu2026infinite} & 3.00 & 24.50 & 15.00 \\
\midrule
\textbf{Ours} & \textbf{20.62} & 41.02 & \textbf{43.67} \\
\bottomrule
\end{tabular}
\end{minipage}
\end{table*}

\begin{figure*}[t]
    \centering
    \includegraphics[width=\linewidth]{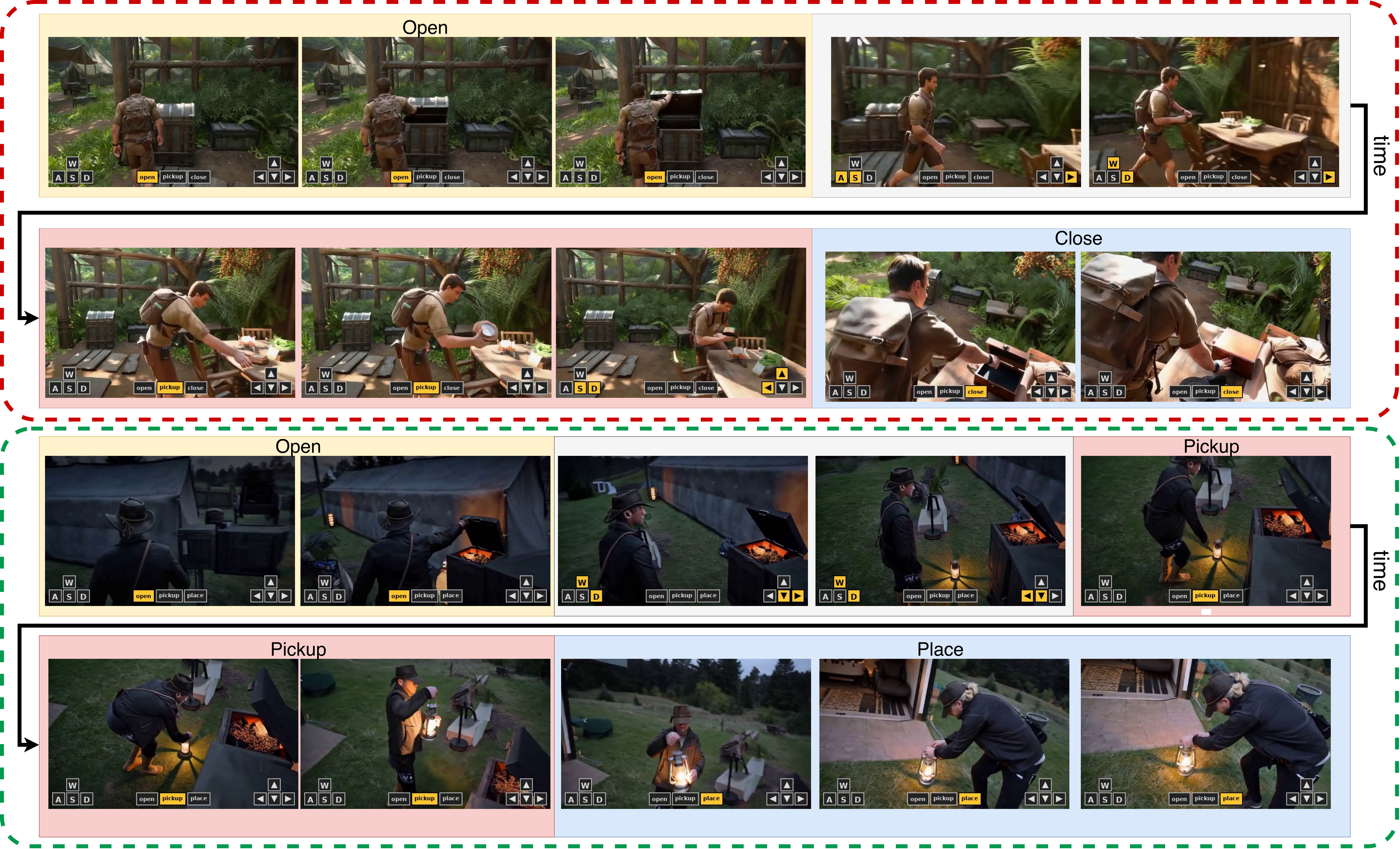}
    \caption{Qualitative visualization of \textbf{\NAME} rollouts across diverse scenes that require both free-form navigation and object-centric interaction within a single continuous trajectory. Each row shows temporally ordered frames with per-frame keyboard/mouse controls overlaid on the images (active inputs highlighted in yellow), illustrating that the model preserves viewpoint controllability while producing coherent interaction outcomes.}
    \label{fig:results_pipeline}
\end{figure*} 

\begin{figure*}[t]
    \centering
    \includegraphics[width=\linewidth]{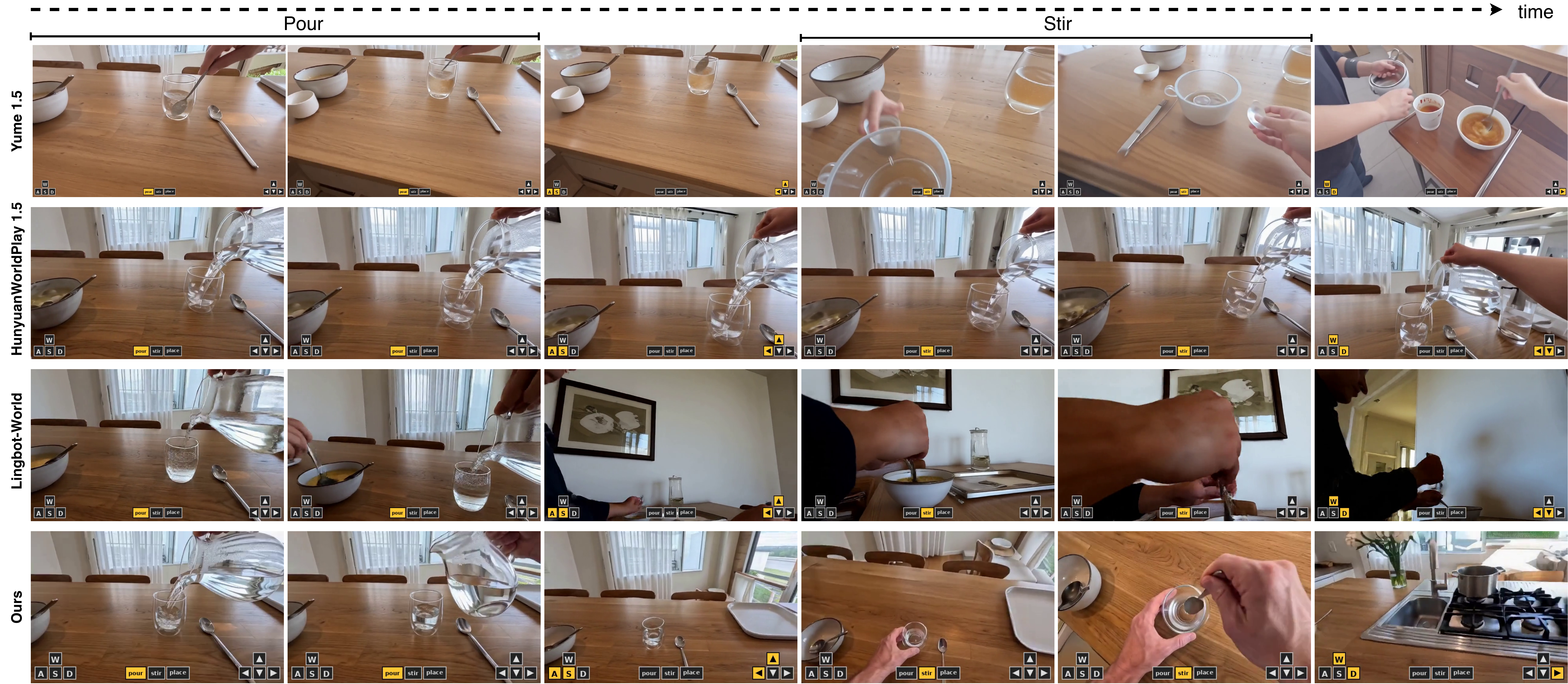}
    \caption{\textbf{Qualitative comparison on a long-horizon, multi-step interaction sequence.} Existing models often fail to do fine-grained human-object interaction or the full manipulation sequence. By contrast, \textbf{\NAME} preserves scene coherence and follows the commanded interaction sequence more faithfully across time.}
    \label{fig:compare}
\end{figure*}

\begin{table*}[t]
\centering
\footnotesize
\renewcommand{\arraystretch}{0.9}   
\setlength{\tabcolsep}{3pt}
\caption{\textbf{Component ablation on I-Bench.} Columns are grouped by evaluation axis: VBench (perceptual/consistency dims; DD/TF/OC dropped due to saturation or low differentiability), VLM-AJ (semantic instruction following), and KMF (geometric controllability). All rows include the Plücker FiLM camera conditioner and per-chunk prompt substitution as baseline components. EAFR: importance-ranked history split; ACHA: action-conditioned history-key amplifier; EventMem: phase-transition tokens.}
\label{tab:ablation}
\begin{tabular}{l | ccccccc | ccc | ccc}
\toprule
\multirow{2}{*}{Run}
& \multicolumn{7}{c|}{\textbf{VBench}} 
& \multicolumn{3}{c|}{\textbf{VLM-AJ}} 
& \multicolumn{3}{c}{\textbf{KMF}} \\
\cmidrule(lr){2-8} \cmidrule(lr){9-11} \cmidrule(lr){12-14}
 & SC & BC & MS & AQ & IQ & i2v-S & i2v-B
 & IF & Succ. & $\geq 2$
 & full & keys & mouse \\
\midrule
CP+Plücker(base)   & 0.803 & 0.895 & 0.985 & 0.479 & 0.714 & 0.934 & 0.942
                     & 2.326 & 52.9 & 80.3
                     & 20.29 & 40.65 & 43.11 \\
+EAFR                & 0.834 & 0.894 & 0.987 & 0.482 & 0.718 & 0.930 & 0.942
                     & 2.345 & 53.8 & 82.1
                     & 20.45 & 40.72 & 43.22 \\
+EAFR+ACHA           & 0.844 & 0.894 & 0.985 & 0.484 & 0.709 & 0.928 & 0.941
                     & 2.413 & 54.0 & 82.8
                     & 20.58 & 40.85 & 43.35 \\
\midrule
\textbf{+EventMem, full}
                     & \textbf{0.871} & \textbf{0.896} & \textbf{0.991} & \textbf{0.485} & \textbf{0.731} & \textbf{0.954} & \textbf{0.957}
                     & \textbf{2.557} & \textbf{57.8} & \textbf{84.5}
                     & \textbf{20.62} & \textbf{41.02} & \textbf{43.67} \\
\bottomrule
\end{tabular}
\end{table*}

\subsection{Evaluation Protocol}
\label{sec:eval_protocol}

We evaluate every ablation along three complementary axes: \textsc{VBench} captures visual quality and temporal consistency, \textsc{VLM-AJ} (VLM-Action-Judge) measures semantic instruction following, and \textsc{KMF} (Key-Mouse-Following) measures geometric controllability of the generated camera trajectory; a faithful interactive world model has to win on all three. All pipelines run on the I-Bench testset introduced in Sec.~\ref{sec:ibench}.

\textbf{(i) VBench perceptual and consistency suite.}
We adopt the same \textsc{VBench-1.0} and \textsc{VBench-i2v} dimensions used by recent video world-model papers~\cite{mao2025yume15,wu2026infinite,lingbot-world}, computed via their native VBench entry points without modification.

\textbf{(ii) VLM-AJ: VLM-Action-Judge.}
VBench is blind to whether the scripted action actually happens; we therefore adopt the four-level instruction-following rubric of WorldModelBench~\cite{Li2025WorldModelBench}, also used by WorldSimBench~\cite{qin2024worldsimbench}. For each chunk a judge VLM is shown four uniformly sampled frames together with the per-chunk GT description and rates action completion on a $0$--$3$ scale (0: action absent; 1: wrong motion; 2: attempted but not completed; 3: fully completed). We report mean score, success rate (Level\,3), and partial-or-better rate (Level\,$\geq 2$), plus a per-action-class breakdown.

\textbf{(iii) KMF: Key-Mouse-Following.}
KMF probes geometric controllability through a closed loop: starting from a ground-truth $(\text{keys},\text{mouse})$ instruction sequence, we generate a video, recover its trajectory, and check whether each chunk moves in the commanded direction. Concretely, given a generated clip we (a)~run VIPE~\cite{huang2025vipe}, the same SLAM-style monocular pose estimator used to label the training corpus, to extract a per-frame $\mathrm{SE}(3)$ trajectory, with frames where SLAM fails recovered by nearest-valid forward/backward fill; (b)~partition the trajectory into the same number of chunks as the GT instruction sequence and uniformly resample each to a $33$-frame window, so that baselines with different native chunk granularities are compared at the same temporal scale; and (c)~map each chunk's pose change to a discrete $(\text{keys},\text{mouse})$ label via simple geometric rules. A predicted chunk is counted as correct only if both its keyboard and mouse labels match the ground truth; the per-video KMF score is the fraction of correct chunks, averaged over all videos in I-Bench. We report joint accuracy $\mathrm{Acc}_{\text{full}}$ together with keys-only and mouse-only accuracies.

\subsection{Results}

We show qualitative results in Fig.~\ref{fig:results_pipeline}. We also compare \NAME against six representative interactive / world-model video generators: Yume~1.5~\cite{mao2025yume15}, HY-World~1.5~\cite{hyworld2025}, Lingbot-World~\cite{lingbot-world}, Matrix-Game~3~\cite{2026matrix}, Astra~\cite{zhu2025astra}, and Infinite-World~\cite{wu2026infinite}. All baselines are evaluated on I-Bench under their official released checkpoints. Tables~\ref{tab:vbench}--\ref{tab:kmf} report the three evaluation axes of Sec.~\ref{sec:eval_protocol}, and Figures~\ref{fig:teaser},~\ref{fig:compare} provide qualitative comparisons. 

\textbf{Quantitative.}
Across the three evaluation axes (Tables~\ref{tab:vbench}--\ref{tab:kmf}), \NAME achieves the best or near-best score on each, with the largest margin on semantic instruction following (Table~\ref{tab:vlm_aj}, where our Level-$3$ success rate of $57.8\%$ more than doubles every baseline). This confirms that the hierarchical action-aware memory addresses the action-forgetting pathology while preserving the visual quality and viewpoint control of navigation-only baselines.

\textbf{Qualitative.}
Figure~\ref{fig:teaser} shows \NAME executing multi-action sequences within a single continuous trajectory under WASD/arrow control, including \textit{Insert}~$\to$~\textit{Pickup} and \textit{Carry}~$\to$~\textit{Place}~$\to$~\textit{Wipe}. More results can be found in Appendix.~\ref{sec:more_vis}. Figure~\ref{fig:compare} contrasts a \textit{Pour}~$\to$~\textit{Stir} sequence in this mixed navigation-and-interaction setting. Existing baselines cannot faithfully follow the fine-grained interaction steps: Yume~1.5 and HY-World~1.5 lose track of the manipulated objects partway through, and Lingbot-World drifts off the scene entirely. This is a systematic limitation of navigation-centric world models, whereas \NAME preserves both the manipulated objects and the commanded action ordering across the full clip. Together, these results validate that navigation and fine-grained object interaction can be unified within a single real-time rollout.

\subsection{Ablation Study}

Table~\ref{tab:ablation} ablates the three memory components on \textsc{I-Bench-mini}. Adding EAFR (re-routing contact and manipulation frames into fine-grained memory) gives a foundational lift; adding ACHA sharpens semantic instruction-following further; Event Memory delivers the largest single jump (subject consistency $0.844 \to 0.871$, Level-3 success $54.0\% \to 57.8\%$), confirming that the persistent event and object slots do the heavy lifting on object identity tracking across long rollouts.

\subsection{User Study}

\begin{wraptable}{r}{0.56\textwidth}
\centering
\footnotesize
\setlength{\tabcolsep}{2pt}
\vspace{-1em}
\caption{\textbf{Results of user study.} Our method outperforms all baselines across all three criteria.}
\label{tab:user_study}
\begin{tabular}{l c c c c c c}
\toprule
 & Astra & HY-W1.5 & LingBot & MG-3 & Yume & \textbf{Ours} \\
\midrule
Action Follow.    & 1.35 & 1.89 & 2.68 & 1.18 & 2.58 & \textbf{4.05} \\
Key/Mouse Follow. & 1.35 & 2.35 & 2.23 & 1.73 & 2.22 & \textbf{3.69} \\
Overall Quality   & 1.31 & 3.15 & 2.62 & 1.70 & 3.02 & \textbf{3.92} \\
\bottomrule
\end{tabular}
\vspace{-1em}
\end{wraptable}

We conduct a user study against the same six baselines on three $1$--$5$ criteria: action following, key/mouse following (overlaid WASD keys and mouse arrows), and overall quality (clarity, temporal consistency, visual artifacts). Tab.~\ref{tab:user_study} shows \NAME ranks first on all three, with the widest margin on action following.

\section{Conclusion}

We presented \textbf{\NAME}, an interactive world model that
extends prior navigation-centric video generators to support
mid-rollout object interaction within a single real-time framework.
Our central argument is that the navigation--interaction gap is
primarily a memory-design issue rather than an architectural one:
recency-based history compression systematically discards the very
frames that drive object-level dynamics.  Our memory and conditioning
innovations address this mismatch and produce a model that performs
object interaction while preserving keyboard-mouse viewpoint control. We view this as a step toward truly interactive world models, with
real-time, AI-generated gameplay as the most immediate downstream
application, and embodied planning and co-driven content creation
following naturally once a learned simulator can both move and act.

\clearpage

\bibliographystyle{plainnat}
\bibliography{main}

\clearpage

\newpage
\appendix
\section*{\huge Appendix}
\newcommand{\appendixhead}%
\appendixhead

\section{Data Generation Pipeline}
\label{sec:supp_data_pipeline}

\begin{figure*}[t]
    \centering
    \includegraphics[width=\linewidth]{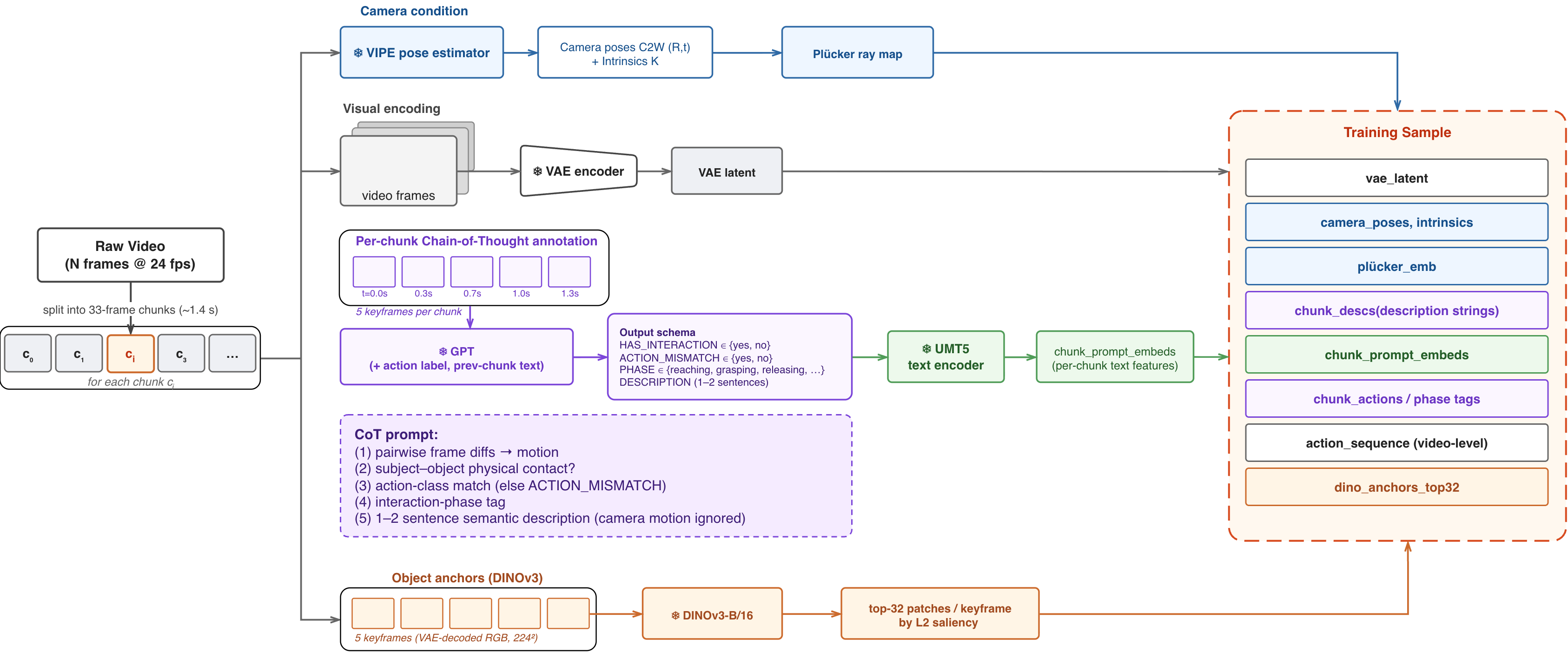}
    \caption{
\textbf{Data generation pipeline.} Each raw 24\,fps video is split into non-overlapping 33-frame chunks ($\approx$1.4\,s). Per chunk, four offline stages run in parallel: VAE-latent extraction (diffusion target); Plücker rays from VIPE-recovered camera poses (FiLM condition for \S\ref{sec:method_kmctrl}); chain-of-thought annotation over five evenly-spaced keyframes via a frozen VLM (GPT-5.4), producing an interaction flag, an action-mismatch flag, a phase label $y^{\text{ph}}_k\in\mathcal{P}$, and a 1--2 sentence description encoded by frozen UMT5; and DINOv3-B/16 object anchors retaining the top-32 saliency-ranked patches per keyframe (candidate pool for the bank writer of \S\ref{sec:method_highlevel_action}). All eight fields form the final cached training sample (right).
}
\label{fig:supp_data_pipeline}
\end{figure*}

This section details the offline data preparation pipeline summarised in Fig.~\ref{fig:supp_data_pipeline}. All stages are run once per video and cached to disk, so no VLM, VAE, or DINOv3 forward pass is required during training; per-step training cost is dominated by the diffusion target alone.

\paragraph{Chunking and visual encoding.}
Each raw 24\,fps video is split into non-overlapping segments of 33 consecutive frames ($\approx$1.4\,s), aligned with the VAE's latent temporal stride. Each chunk is encoded by a frozen VAE into a latent tensor that serves as the diffusion target during training and the autoregressive output during rollout. Chunks for which any of the 33 source frames is invalid (scene cut or missing frame) are dropped together with their downstream annotations.

\paragraph{Camera condition.}
Per-frame camera-to-world poses $(\mathbf{R}, \mathbf{t})$ and intrinsics $\mathbf{K}$ are recovered with the VIPE pose estimator~\citep{huang2025vipe}. From $(\mathbf{R}, \mathbf{t}, \mathbf{K})$ we build a per-pixel Plücker-ray map and downsample it to the latent spatial grid; this map drives the PlückerFiLM modulation of \S\ref{sec:method_kmctrl} (full construction in \S\ref{sec:supp_kmctrl}). Poses are normalised to the chunk's first frame so the conditioning is translation-invariant within the chunk.

\paragraph{Per-chunk chain-of-thought annotation.}
For each chunk we extract five evenly-spaced keyframes ($t\!\in\!\{0.0, 0.3, 0.7, 1.0, 1.3\}$\,s) and query a frozen vision-language model (GPT-5.4) under a chain-of-thought prompt that has access to (i) the five keyframes, (ii) the dataset-supplied video-level action label, and (iii) the description of the previous chunk for temporal continuity. The reasoning template walks the model through five steps:

\begin{enumerate}\setlength{\itemsep}{0pt}
    \item compute pairwise frame differences and summarise the dominant motion;
    \item determine whether subject and object are in physical contact;
    \item check whether the supplied action label is consistent with the observed motion (otherwise emit \texttt{ACTION\_MISMATCH});
    \item assign an interaction-phase tag $y^{\text{ph}}_k\in\mathcal{P}$ from the taxonomy of \S\ref{sec:cot_caption};
    \item write a 1--2 sentence semantic description grounded solely in the per-frame evidence and ignoring camera motion.
\end{enumerate}

The structured output schema is fixed: \texttt{HAS\_INTERACTION} $\in$ \{yes, no\} (the interaction flag $y^{\text{int}}_k$), \texttt{ACTION\_MISMATCH} $\in$ \{yes, no\}, \texttt{PHASE} $\in\mathcal{P}$ (the phase label $y^{\text{ph}}_k$), plus a free-form \texttt{DESCRIPTION} string. Together with the video-level action class $a\!\in\!\mathcal{A}$ from the dataset manifest (Appendix.~\ref{sec:supp_actions}), these form the per-chunk label triple $(y^{\text{int}}_k, y^{\text{ph}}_k, a)$ consumed by the three memory modules of \S\ref{sec:method_highlevel_action}.

\paragraph{Object anchors via DINOv3.}
For each of the same five keyframes we run a frozen DINOv3-B/16 encoder on a $224\!\times\!224$ centre crop and retain the 32 patches with the largest $L_2$ feature norm per keyframe as candidate object anchors. To keep the saliency distribution consistent with inference, where the bank writer only sees model-generated content, the DINOv3 input frames are decoded from the chunk's VAE latent rather than read from raw RGB; this train--test alignment removes the small but non-trivial domain gap introduced by VAE compression. The pre-baked top-32 patches form the candidate pool from which the action-aware memory bank's writer selects at training time (full wiring in \S\ref{sec:supp_bank_impl}).

\paragraph{Final cached training sample.}
Each cached sample bundles eight fields:
\begin{itemize}\setlength{\itemsep}{0pt}
    \item \texttt{vae\_latent}: VAE-encoded 33-frame chunk (diffusion target);
    \item \texttt{camera\_poses}, \texttt{intrinsics}: per-frame $(\mathbf{R}, \mathbf{t})$ and $\mathbf{K}$;
    \item \texttt{plucker\_emb}: pre-rendered Plücker-ray map at latent resolution;
    \item \texttt{chunk\_descs}: raw per-chunk description string (diagnostics only);
    \item \texttt{chunk\_prompt\_embeds}: UMT5 embedding of the description ($\mathbf{e}^{\mathrm{chunk}}_t$);
    \item \texttt{chunk\_actions}: structured labels $(y^{\text{int}}_k, y^{\text{ph}}_k, \texttt{ACTION\_MISMATCH})$ per chunk;
    \item \texttt{action\_sequence}: video-level action class $a$ from the 40-verb vocabulary (Appendix.~\ref{sec:supp_actions});
    \item \texttt{dino\_anchors\_top32}: pre-baked DINOv3 top-32 patches per keyframe (5 keyframes $\times$ 32 patches).
\end{itemize}
All fields are written to disk once per dataset; the runtime data loader concatenates them into the model-side inputs without further processing.

\section{Label Handling and Zero-Initialisation Invariants}
\label{sec:supp_labels}

\paragraph{Fall-back on unannotated clips.}
Clips that lack one or more of the per-chunk labels $(y^{\text{int}}_k,\,y^{\text{ph}}_k,\,a)$ defined in \S\ref{sec:supp_data_pipeline} (e.g., pure navigation data scraped without an interaction track) emit \texttt{None} for the missing fields. EAFR's frame importance score defaults to its recency component only when $y^{\text{ph}}$ is absent, ACHA's MLP receives a zero action embedding when $a$ is absent, and the action-aware memory bank's writer gate fires only when the relevant labels are present. In all three cases the per-sample fallback is bit-for-bit equivalent to the baseline, so mixed batches of annotated and unannotated clips train without any distributional shift.

\subsection{Zero-initialisation Invariants}
\label{sec:supp_zero_init}

A subtle pitfall when adding modules to a pretrained backbone is that the mere presence of a new module at step~0 can shift the forward pass---through a non-zero output projection, a RoPE re-indexing, or a moved token---making any downstream metric change ambiguous between ``the module learned something useful'' and ``the model recovered from a worse initialisation''. Our design neutralises this on every new path:

\begin{itemize}\setlength{\itemsep}{0pt}
\item \textbf{EAFR.} Frames are re-assigned across short/mid/long buckets but their RoPE indices follow the original time stamps, so self-attention sees the same temporal embedding it would in the time-only baseline; the only thing that changes is which compression kernel each frame meets.
\item \textbf{ACHA.} The bottleneck MLP $W_2\,\sigma(W_1\,\mathbf{e}_a)$ that produces the action-conditioned delta on top of the per-head amplifier $s$ has its final layer zero-initialised, so $\alpha(\mathbf{e}_a)\!=\!\operatorname{softplus}(s)$ at step~0, recovering the unconditioned amplifier.
\item \textbf{Plücker FiLM.} The scale and shift heads $W_s,W_b$ in \eqref{eq:plucker_film} are zero-initialised; combined with frame-0 pose normalisation this guarantees $\widetilde{\mathbf{X}}^{(\ell)}_{\text{cur}}=\mathbf{X}^{(\ell)}_{\text{cur}}$ at step~0, so attaching the branch to a trained checkpoint does not disturb the existing forward pass.
\item \textbf{Action-aware memory bank.} The bank's segment embedding, the event token's out-projection, and the object token's final visual projection $W_v$ are all zero-initialised, so every prepended bank token contributes exactly $\mathbf{0}$ to the self-attention sum until trained.
\end{itemize}

Together these invariants ensure that toggling any subset of the new modules on a pretrained checkpoint produces a step-0 loss curve that matches the baseline to machine precision; every subsequent metric change is therefore attributable to learning, not to a re-init shock.

\section{Keyboard/Mouse Control Conditioning: implementation details}
\label{sec:supp_kmctrl}

This subsection expands on the geometric Plücker branch and the symbolic text-camera branch summarised in \S\ref{sec:method_kmctrl}.

\paragraph{Plücker-ray construction.}
Each training clip ships with a per-frame camera-to-world pose $\mathbf{P}_t\!\in\!\mathrm{SE}(3)$ and intrinsics $\mathbf{K}\!=\![f_x, f_y, c_x, c_y]$ recovered by VIPE~\citep{huang2025vipe}. We normalise translations relative to frame~0 (rotations are unchanged), then downsample to one pose per latent frame ($T_p{=}33$ video frames $\to T_l{=}9$ latent frames). For each latent frame $t$ and each pixel $(u,v)$ on the latent grid, we cast a ray from the camera origin and form the 6-D Plücker representation
\begin{equation}
    \boldsymbol{\rho}_{t,u,v} \;=\; \bigl[\,\mathbf{r}_o(t)\!\times\!\mathbf{r}_d(t,u,v),\; \mathbf{r}_d(t,u,v)\,\bigr] \in \mathbb{R}^{6},
\end{equation}
where $\mathbf{r}_o(t)$ is the world-space camera origin at frame~$t$ and $\mathbf{r}_d(t,u,v)$ is the world-space ray direction for pixel $(u,v)$ at frame~$t$. This yields a tensor $\boldsymbol{\rho}\!\in\!\mathbb{R}^{B\times 6\times T_l\times H_p\times W_p}$ aligned one-to-one with the latent grid.

\paragraph{Patch packing ($\mathrm{Pack}$).}
The DiT operates on latent patch tokens. The packing operator $\mathrm{Pack}$ rearranges the 6-D Plücker rays inside each patch into a single feature vector, producing one packed block per latent patch-token. The subsequent linear $W_{\text{patch}}$ projects each packed block into the transformer width $d$.

\paragraph{PlückerFiLM module.}
Given $\boldsymbol{\rho}$, the module first produces a per-patch cam token via a residual MLP:
\begin{equation}
    \mathbf{z}^{\text{cam}} \;=\; W_2\,\sigma\!\bigl(W_1\,\mathbf{e}\bigr) + \mathbf{e},
    \qquad
    \mathbf{e} \;=\; W_{\text{patch}}\!\bigl(\mathrm{Pack}(\boldsymbol{\rho})\bigr),
    \label{eq:plucker_encode}
\end{equation}
where $\sigma$ is SiLU. At every transformer block~$\ell$, the same $\mathbf{z}^{\text{cam}}$ drives a per-token FiLM scale-and-shift modulation applied only to the current-chunk hidden states $\mathbf{X}^{(\ell)}_{\text{cur}}$ (history tokens from the KV cache are untouched):
\begin{align}
    \mathbf{c} &= W_4\,\sigma\!\bigl(W_3\,\mathbf{z}^{\text{cam}}\bigr) + \mathbf{z}^{\text{cam}},\nonumber\\
    \widetilde{\mathbf{X}}^{(\ell)}_{\text{cur}}
    &= \bigl(\mathbf{1}+\mathbf{s}\bigr)\odot\mathbf{X}^{(\ell)}_{\text{cur}} + \mathbf{b},
    \quad
    \mathbf{s} = W_{s}\,\mathbf{c},\;\;
    \mathbf{b} = W_{b}\,\mathbf{c}.
    \label{eq:plucker_film}
\end{align}
The scale/shift heads $W_s, W_b$ are zero-initialised; combined with the frame-0 pose normalisation, this gives a step-0 identity $\widetilde{\mathbf{X}}^{(\ell)}_{\text{cur}} = \mathbf{X}^{(\ell)}_{\text{cur}}$, so enabling the branch on a pretrained checkpoint preserves the baseline forward pass exactly. The same module weights $\{W_1,\dots,W_4, W_s, W_b\}$ are reused at every transformer block, contributing $\approx$1.2\% of the 14\,B DiT.

\paragraph{Per-token vs.\ broadcast modulation.}
The FiLM signal in Eq.~\ref{eq:plucker_film} is per-token rather than broadcast over space, since yaw/pitch rotations and translations move different image regions differently---e.g., a camera yaw shifts the right side of the image faster than the left in pixel velocity. A single global control vector cannot express this anisotropy.

\paragraph{Symbolic command vocabulary.}
The dataset is annotated with $9{\times}9{=}81$ \texttt{(keyboard, mouse)} combinations: 9 keyboard categories $\{$\textit{idle}, W, A, S, D, W+A, W+D, S+A, S+D$\}$ crossed with 9 mouse categories $\{$\textit{idle}, $\uparrow$, $\downarrow$, $\leftarrow$, $\rightarrow$, $\uparrow\!\rightarrow$, $\uparrow\!\leftarrow$, $\downarrow\!\rightarrow$, $\downarrow\!\leftarrow\}$. Each combination is mapped to a short natural-language template of the form ``Person moves $\langle$keyboard direction$\rangle$. Camera tilts $\langle$mouse direction$\rangle$.'' Examples:
\begin{itemize}\setlength{\itemsep}{0pt}\setlength{\parskip}{0pt}
    \item \texttt{(idle, idle)}: ``Person stays still. Camera holds steady.''
    \item \texttt{(W, $\uparrow\!\rightarrow$)}: ``Person moves forward. Camera tilts up and turns right.''
    \item \texttt{(W{+}D, $\uparrow\!\rightarrow$)}: ``Person moves forward and right. Camera tilts up and turns right.''
\end{itemize}

\paragraph{Cross-attention conditioning.}
Each command template is encoded once by the frozen UMT5 text encoder, producing an embedding $\mathbf{e}^{\text{cam-txt}}\!\in\!\mathbb{R}^{L_a\times d_t}$ that we cache so neither training nor inference pays a text-encoder cost. At training time the per-chunk action label indexes into this cache; the embedding is concatenated with the per-chunk caption embedding from \S\ref{sec:cot_caption} along the sequence dimension before the DiT cross-attention:
\begin{equation}
    \mathbf{h}_{\text{enc}}
    \;=\;
    \bigl[\,\mathbf{e}^{\text{chunk}}_t\,;\;\mathbf{e}^{\text{cam-txt}}_t\,\bigr]
    \;\in\;\mathbb{R}^{(L+L_a)\times d_t}.
    \label{eq:cam_text_concat}
\end{equation}
Because the symbolic label is a tempting shortcut---one short phrase shared across a whole bucket of trajectories---we apply an independent dropout $p_{\text{cam-txt}}{=}0.1$ to $\mathbf{e}^{\text{cam-txt}}_t$ before concatenation, leaving $\mathbf{e}^{\text{chunk}}_t$ and the geometric Plücker tensor $\boldsymbol{\rho}$ intact. This biases the model toward the geometric branch and, empirically, drives the zero-initialised FiLM heads off the identity manifold.

\section{Action-aware Memory Bank: implementation details}
\label{sec:supp_bank_impl}

\paragraph{Hyperparameters.}
We use $K_{\text{tot}}{=}16$ for total bank capacity, $K_{\text{kf}}{=}5$
keyframes per chunk for the object writer's DINOv3 forward, and
$K_{\text{pc}}{=}3$ object-anchor tokens kept per qualifying chunk.
The event writer fires at most once per chunk per transition tag
($\le 3$ events / chunk in practice). The object writer's auxiliary
trigger phases (used in addition to $y^{\text{int}}_k\!=\!1$) are
$\{\textit{completing}, \textit{post-action}\}$.

\paragraph{Bank insertion and attention path.}
At every chunk-autoregressive step, the bank's contents are prepended
to the DiT input sequence ahead of the current-chunk noise tokens.
Each slot carries a learnable segment embedding with one extra bit
distinguishing event from object tokens, so the DiT layers can
specialise to token type without recovering it from content alone.
Bank tokens are written with all RoPE positional indices set to zero,
so they participate in attention purely through their content
embeddings without competing with the spatial RoPE grid on
current-chunk patches. They enter the DiT self-attention only and do
not modify the cross-attention path, which continues to carry the
per-chunk semantic embedding $\mathbf{e}^{\text{chunk}}_t$ from
\S\ref{sec:cot_caption}. With $K_{\text{tot}}$ bank tokens and
current-chunk latent token counts on the order of
$F\!\cdot\!H\!\cdot\!W$ (thousands), the bank adds well under
$0.5\%$ to the DiT input length and a comparable fraction to per-step
compute.

\paragraph{FIFO and pinning.}
The bank is a fixed-capacity deque of size $K_{\text{tot}}$; when
full, the oldest unpinned token is evicted. Tokens from chunks labelled
\textit{contact}, \textit{manipulating}, or \textit{completing} are pinned
for the remainder of the rollout, while tokens from \textit{approaching},
\textit{reaching}, or \textit{post-action} chunks recycle normally.

\paragraph{DINOv3 backbone.}
We use the publicly released DINOv3-B/16 (ViT-B with $14\!\times\!14$
patch tokens at $224^2$ input, $768$-dim outputs). Inputs are
bilinearly resized from the chunk's native $384\!\times\!640$
keyframes and normalised with the standard ImageNet mean/std. The
$L_2$ norm of each patch feature serves as the saliency score; we
flatten the $K_{\text{kf}}\!\times\!P$ patches into a single pool and
take the top $K_{\text{pc}}$ across the chunk so a salient object
caught in two consecutive keyframes contributes both views to the
bank instead of being deduplicated.

\paragraph{Offline pre-baking.}
For every training video we run DINOv3 once over its keyframes and
cache the top-32 patch features and saliency scores per chunk
(\texttt{dino\_anchors\_top32}, $\approx 50$\,KB / chunk in fp16) into
the feature \texttt{.pt} file. The training-time writer reads the
cache and selects the top $K_{\text{pc}}$ per chunk; this pushes the
DINOv3 forward off the training critical path entirely. Pre-baking
the full $\sim$100\,K-clip set takes $\sim$2.5\,h on
$8\!\times\!$H100 (one rank per video shard).

\paragraph{Online inference.}
At inference the model produces brand-new chunks that have no
pre-baked features. The pipeline therefore (i) VAE-decodes the
just-generated chunk, (ii) samples $K_{\text{kf}}$ evenly-spaced
keyframes from the decoded $33$-frame block, (iii) runs a single
forward through frozen DINOv3 to obtain patch features, and
(iv) admits the top patches under the same writer gate
($y^{\text{int}}_k\!=\!1$ or
$y^{\text{ph}}_k\!\in\!\{\textit{completing}, \textit{post-action}\}$).
Pure-navigation chunks skip the decode + DINOv3 pass entirely. In
practice $\sim$30--40\,\% of chunks fire the writer; the average
inference overhead is below $15\,\%$ of the chunk-generation budget.
Event tokens require no online computation: they operate purely on
the symbolic $y^{\text{ph}}$ stream the model already conditions on,
and the embedding lookup
$E_\xi[\xi_k] + E_{\text{ph}}[y^{\text{ph}}_k] + E_a[a]$ is
constant-time.

\section{Distillation Details}
\label{sec:supp_distill}

This section provides full technical details of the distillation pipeline summarised in \S\ref{sec:method_distill}. The recipe closely follows Helios~\cite{yuan2026helios}; we restate the relevant equations here for self-containment.

\paragraph{Stage 2: Multi-resolution flow matching.}
We split denoising into $K{=}3$ resolution levels, partition $[0,1000]$ into boundaries $T_0{=}1000 {>} T_1 {>} T_2 {>} T_3{=}0$, and at each level $k$ train along the linear flow path with ground-truth velocity $\mathbf{v}^{k}=\mathbf{x}^{k}-\mathrm{Upsample}(\mathbf{x}^{k-1})$,
\begin{equation}
    \mathcal{L}_{\text{flow}}
    \;=\; \mathbb{E}\bigl[\,\lVert
        u^{k}_{\theta}(\mathbf{x}^{k}_{t},\mathbf{y},\lambda_{t},k)
        - \mathbf{v}^{k}\rVert_{2}^{2}\,\bigr],
    \label{eq:supp_flow}
\end{equation}
where $\mathbf{y}$ bundles the chunk caption, the text-camera embedding, the action label, and the Plücker tensor. At inference we allocate $(N_1,N_2,N_3)$ steps across the three levels and bridge transitions by nearest-neighbour upsampling followed by re-noising; coarser levels process fewer tokens, so the per-chunk cost drops substantially.

\begin{figure*}[t!]
    \centering
    \includegraphics[width=0.96\linewidth]{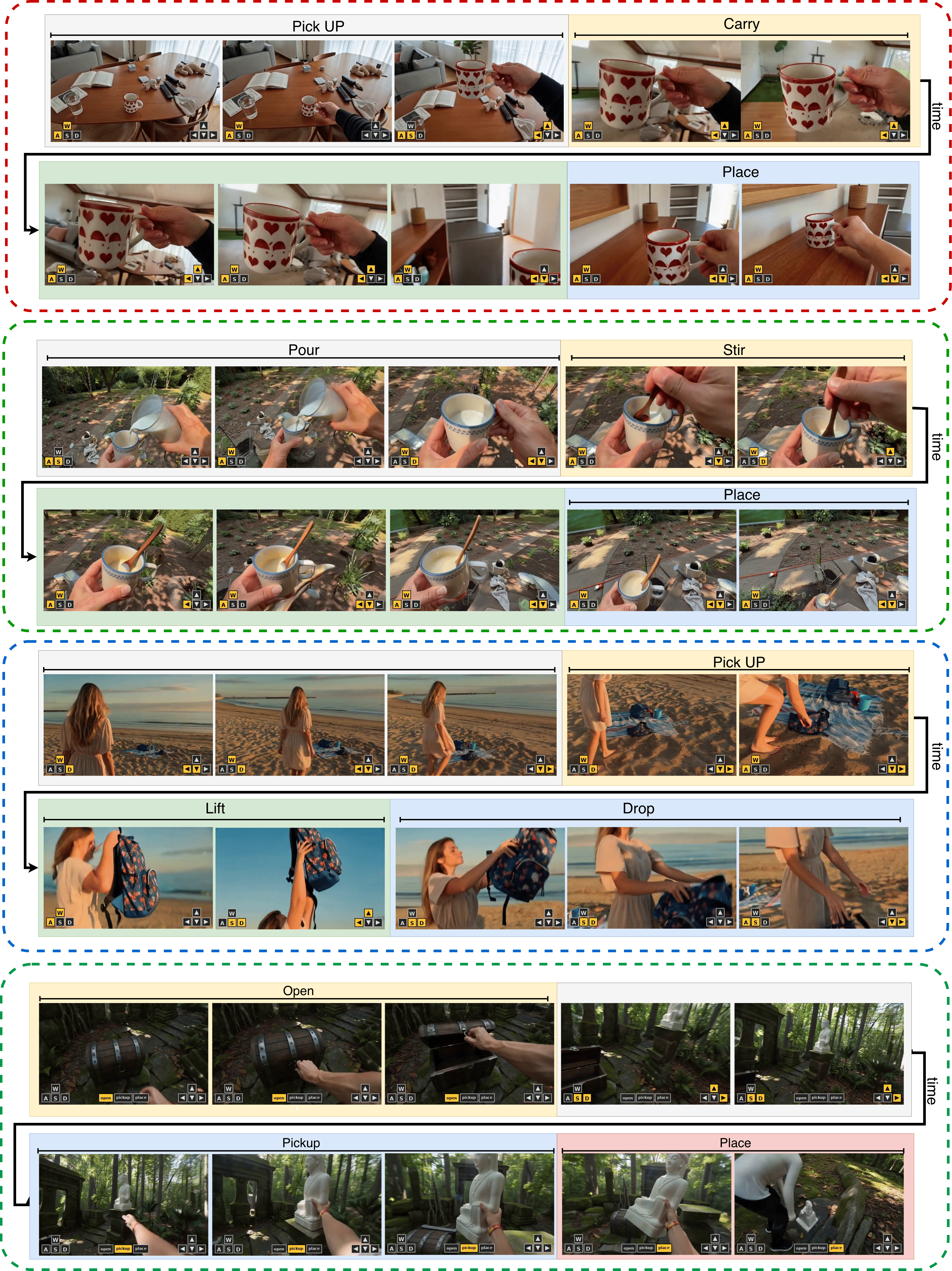}
    \caption{Further results generated by ActWorld.}
    \label{fig:vis_supp1}
\end{figure*}

\paragraph{Stage 3: Backward simulation under pure teacher forcing.}
We distill the 50-step model into a 3-step generator $G_\theta$ with a DMD-style~\cite{yin2024improved,yin2024onestep} objective tailored to the autoregressive chunk setting. Training uses pure teacher forcing: only real history latents are fed, and we generate a single chunk per step rather than rolling out long sequences. Backward simulation is performed in $K$ levels matching the flow-matching pyramid; at level $k$ we estimate
\begin{equation}
    \mathbf{x}^{k}_{0}
    \;=\; \mathbf{x}^{k}_{t}
        - \lambda_{t}\,u^{k}_{\theta}(\mathbf{x}^{k}_{t},\mathbf{y},\lambda_{t},k),
    \label{eq:supp_xk0}
\end{equation}
reconstruct $\mathbf{x}^{k}_{t}$ along the flow path, iterate to convergence, and pass $\mathbf{x}^{k}_{0}$ to level $k{+}1$. The I2V ODE warm-up used to initialize $G_\theta$ before Stage 3 begins is described in \S\ref{sec:method_distill} of the main paper and is the only deviation from~\cite{yuan2026helios}.

\paragraph{Stage 3 objectives.}
Distribution matching uses a CFG'd real score $p_{\text{real}}$ and a fake score $p_{\text{fake}}$. To escape the teacher's expressive ceiling we attach multi-granularity discriminator heads $D$ to selected DiT layers of $p_{\text{fake}}$ and train them on real data with a non-saturated GAN loss plus an approximate-$R_1$ regulariser~\cite{lin2025diffusion}. The Stage-3 objectives are
\begin{align}
    \mathcal{L}_{G_\theta}        &= \mathcal{L}_{\text{DMD}}  + w_G\,\mathcal{L}_G,
    \label{eq:supp_lg}\\
    \mathcal{L}_{p_{\text{fake}}} &= \mathcal{L}_{\text{Flow}} + w_D\,\mathcal{L}_D,
    \label{eq:supp_lpfake}
\end{align}
and $G_\theta$ is updated once per several $p_{\text{fake}}$ updates following the TTUR scheduling of~\cite{yin2025causvid}.

\section{More Visualizations}
\label{sec:more_vis}

We show further visualization in Fig.~\ref{fig:vis_supp1}.

\section{Training Details}
\label{sec:supp_training}

\paragraph{Initialisation.}
We initialise from a publicly released Helios~\cite{yuan2026helios} 14\,B chunk-autoregressive checkpoint, itself adapted from the Wan2.1-14B~\cite{wan2025} bidirectional text-to-video diffusion model pretrained on open-domain data. All architectural additions of \S\ref{sec:method_kmctrl}--\ref{sec:method_highlevel_action} (Plücker FiLM, ACHA bottleneck, memory-bank embedding tables, object-token MLP $W_v$) have zero-initialised final layers, so at step~0 the modified checkpoint is bit-for-bit equivalent to Helios.

\paragraph{Task.}
Stage-1 training is run in an image-to-video (I2V) regime: for each clip the first ground-truth chunk is fixed as a clean conditioning image, and the model is supervised to denoise subsequent chunks under the per-chunk caption (\S\ref{sec:cot_caption}), Plücker rays and symbolic camera command (\S\ref{sec:method_kmctrl}), and action-aware memory channel (\S\ref{sec:method_highlevel_action}). The same I2V interface is preserved at inference, so the deployed model always conditions on a user-provided starting frame.

\paragraph{Hardware and schedule.}
Stage-1 trains for 5{,}000 optimisation steps on $48\!\times\!$H100 GPUs with bf16 mixed precision. Stage-2 multi-resolution flow-matching warm-up trains for another 5{,}000 steps on the same hardware. Stage-3 distillation comprises a 3{,}000-step I2V ODE warm-up followed by 3{,}000 steps of GAN distillation. Both stages follow the Helios~\cite{yuan2026helios} schedule; full equations and the I2V ODE-pair construction are deferred to Appendix.~\ref{sec:supp_distill}.

\section{High-Level Action Vocabulary}
\label{sec:supp_actions}

Our training corpus is annotated with 40 high-level action commands selected to span both human-centric manipulation and dynamic locomotion. We organise them into eight semantic groups:

\begin{itemize}\setlength{\itemsep}{2pt}
    \item \textbf{Pick-and-place (8):} \texttt{pickup}, \texttt{putdown}, \texttt{place}, \texttt{lift}, \texttt{drop}, \texttt{hold}, \texttt{grab}, \texttt{carry}.
    \item \textbf{Open/close \& lock (4):} \texttt{open}, \texttt{close}, \texttt{lock}, \texttt{unlock}.
    \item \textbf{Attach/detach (6):} \texttt{attach}, \texttt{detach}, \texttt{insert}, \texttt{remove}, \texttt{plug}, \texttt{unplug}.
    \item \textbf{Force application \& locomotion (6):} \texttt{push}, \texttt{pull}, \texttt{drag}, \texttt{throw}, \texttt{kick}, \texttt{drive}.
    \item \textbf{Surface and tool manipulation (6):} \texttt{wipe}, \texttt{swipe}, \texttt{stir}, \texttt{pour}, \texttt{peel}, \texttt{cut}.
    \item \textbf{Discrete contact (3):} \texttt{tap}, \texttt{press}, \texttt{switch}.
    \item \textbf{Handover (3):} \texttt{give}, \texttt{receive}, \texttt{release}.
    \item \textbf{Orientation \& material (4):} \texttt{rotate}, \texttt{slide}, \texttt{fold}, \texttt{pack}.
\end{itemize}

The corpus is partitioned across three sources by viewpoint and subject: a 50K-clip first-person ego-centric subset spanning all 40 categories (1{,}250 clips each); a 30K-clip third-person human-manipulation subset covering the 32 manipulation-focused categories ($\sim$938 each); and a 10K-clip third-person non-human subset (animals, vehicles) restricted to the 8 dynamic-motion categories \texttt{carry}, \texttt{drag}, \texttt{drive}, \texttt{drop}, \texttt{grab}, \texttt{hold}, \texttt{pull}, \texttt{release} (1{,}250 each), totaling 90{,}000 annotated clips.

\end{document}